%% file: parallaxrag_acl.tex
\documentclass[11pt]{article}

\usepackage[final]{acl}

\usepackage{times}
\usepackage{latexsym}
\usepackage{caption}
\usepackage[T1]{fontenc}

\usepackage[utf8]{inputenc}

\usepackage{microtype}

\usepackage{inconsolata}

\usepackage{graphicx}

\usepackage{url}
\usepackage{booktabs}
\usepackage{multirow}
\usepackage{makecell}
\usepackage{enumitem} 
\usepackage{subcaption}
\usepackage{wrapfig}
\usepackage{algorithm}
\usepackage{algorithmicx}
\usepackage{tabularx}   
\usepackage[noend]{algpseudocode}
\usepackage{float}

\usepackage[breakable,skins]{tcolorbox}
\usepackage{courier} 
\input{ParallaxRag-math_commands.tex}

\title{Think Parallax: Solving Multi-Hop Problems via Multi-View Knowledge-Graph-Based Retrieval-Augmented Generation}

\author{
  Jinliang Liu$^{1,2}$,
  Jiale Bai$^{2}$,
  Shaoning Zeng$^{1,2}$\thanks{Corresponding author} \\
  $^{1}$Yangtze Delta Region Institute (Huzhou), University of Electronic Science and Technology of China,\\
  Huzhou, Zhejiang, China \\
  $^{2}$School of Information and Software Engineering, University of Electronic Science and Technology of China, \\
  Chengdu, Sichuan, China \\
  \texttt{lucaliu510@gmail.com, syxb02@gmail.com, zsn@outlook.com} \\
}

\begin{document}
\maketitle

\begin{abstract}
Large language models (LLMs) still struggle with multi-hop reasoning over knowledge-graphs (KGs), and we identify a previously overlooked structural reason for this difficulty: Transformer attention heads naturally specialize in distinct semantic relations across reasoning stages, forming a hop-aligned relay pattern. This key finding suggests that multi-hop reasoning is inherently multi-view, yet existing KG-based retrieval-augmented generation (KG-RAG) systems collapse all reasoning hops into a single representation, flat embedding space, suppressing this implicit structure and causing noisy or drifted path exploration. We introduce ParallaxRAG, a symmetric multi-view framework that decouples queries and KGs into aligned, head-specific semantic spaces. By enforcing relational diversity across multiple heads while constraining weakly related paths, ParallaxRAG constructs more accurate, cleaner subgraphs and guides LLMs through grounded, hop-wise reasoning. On WebQSP and CWQ, it achieves state-of-the-art retrieval and QA performance, substantially reduces hallucination, and generalizes strongly to the biomedical BioASQ benchmark. Our implementation is available at \url{https://github.com/LucaLiu1313/ParallaxRAG}.
\end{abstract}

\begin{figure}[h]
    \centering
    \includegraphics[width=1\columnwidth]{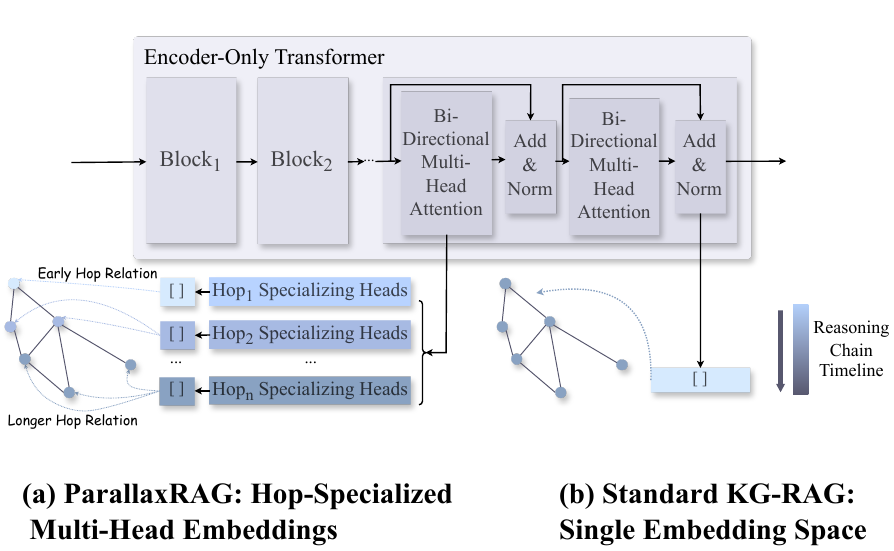}
    \caption{Comparison of ParallaxRAG and standard KG-RAG embedding strategies. (a) In ParallaxRAG, attention heads specialize in distinct semantic relations across reasoning stages, where early heads capture shallow relational patterns while later heads handle longer-hop dependencies. (b) In standard KG-RAG, all reasoning hops are collapsed into a single embedding space, suppressing hop-specific semantic structure.}
    \label{fig:transformer}
    \vspace{-10pt}
\end{figure}

\section{Introduction}
Multi-hop reasoning over KGs remains a challenge for LLMs. 
Although RAG improves factual grounding~\cite{lewis2020retrieval}, existing KG-RAG systems still struggle with long reasoning chains~\cite{luo2023reasoning}, compounding errors~\cite{mavromatis2024gnn}, and the rapid expansion of irrelevant graph paths~\cite{he2024g}. 
These issues limit both the accuracy and robustness of downstream question answering.

Our analysis identifies an architectural property of Transformers that is directly relevant to this problem, where different attention heads consistently specialize in distinct semantic relations across reasoning stages, forming a hop-aligned pattern. 
Some attention heads primarily contribute to the early stages of reasoning, while others specialize in later-stage reasoning and make dominant contributions as the reasoning process deepens.
This observation suggests that multi-hop reasoning is inherently multi-view, with different hops requiring distinct representational subspaces.

Current KG-RAG approaches do not leverage this structure.  
Most methods encode queries and graph triples into a single embedding space~\cite{fu2020survey, ji2021survey}, which conflates hop-specific semantics and hinders step-wise reasoning.  
This monolithic representation causes early-hop signals to interfere with deeper relational composition, increasing noise and reducing retrieval precision~\cite{liu2023lost}.  
Iterative LLM-based reasoning~\cite{sun2023think} offers partial improvements but introduces high latency and lacks an explicit mechanism for hop decomposition over KGs.  
These limitations indicate a structural mismatch: multi-hop reasoning requires hop-separated semantics, yet existing KG-RAG systems enforce a single shared embedding basis.

To instantiate this principle, we introduce ParallaxRAG, a framework based on symmetric multi-head decoupling. We leverage transformer multi-head activations to decompose queries into multi-view representations~\cite{vaswani2017attention, besta2024multi}, while projecting the KG triples into aligned multi-faceted latent spaces~\cite{mavromatis2024gnn, li2024simple}. Two mechanisms operationalize this framework: (1) a Pairwise Similarity Regularization (PSR) module integrated into the Distance Encoding (DE) stage, which enforces head-level specialization and prevents representational collapse; and (2) a lightweight retrieval component that consolidates head-specific information to reduce noise and improve alignment between queries and graph structure~\cite{khattab2020colbert}.

Our main contributions are as follows:
\begin{itemize}
\item \textbf{Multi-head decoupling architecture for KG-RAG.}  
We propose the first KG-RAG framework that decouples queries and KGs into head-specific views, allowing attention heads to capture complementary relational cues at different reasoning depths. We also introduce a joint exploration–exploitation strategy, namely PSR and weakly supervised gating which enhances robustness and promotes specialization.

\item \textbf{Head specialization in multi-hop reasoning.}  
We provide the first empirical evidence of a relay effect, where distinct head groups dominate different stages of multi-hop reasoning. New metrics quantify head-level contribution and effectiveness.

\item \textbf{State-of-the-Art Performance and Cross-Domain Generalization.}
ParallaxRAG achieves state-of-the-art results on WebQSP and CWQ, and generalizes to the biomedical BioASQ benchmark. Under zero-shot transfer, it surpasses the previous SOTA by 7.32 Macro-F1, demonstrating transferable reasoning capabilities across domains.
\end{itemize}

\begin{figure*}[!t]
    \centering
    \includegraphics[width=1\textwidth]{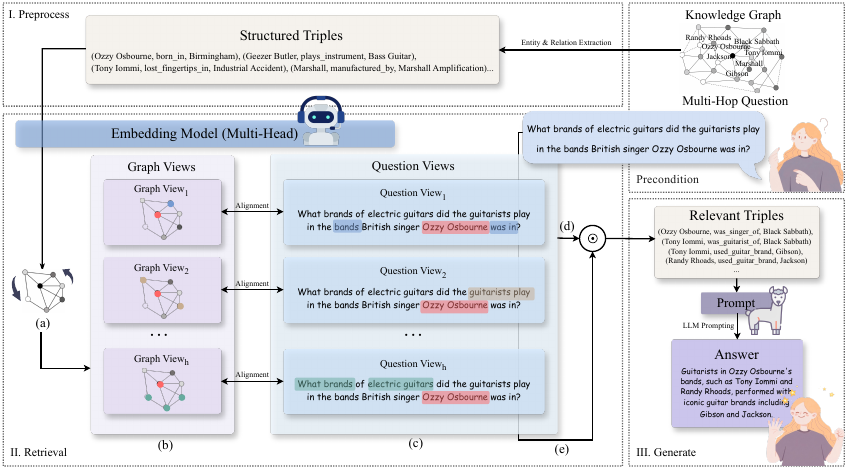}
    \caption{The architecture of ParallaxRAG, illustrating its three main stages: Preprocess, Retrieval, and Generate. The core retrieval process begins by creating specialized multi-view representations for both the graph (b) and the question (c), a process guided by directional distance encoding and regularized by pairwise similarity (a). Next, candidate triples are scored in parallel across these diverse views (d). Finally, a query-aware gating mechanism computes head importance (e) to produce a weighted aggregation of the scores, identifying the most relevant triples for the final generation stage.}
    \label{fig:example2}
\end{figure*}

\section{Related Work} \label{headings}

\subsection{Retrieval-Augmented Generation (RAG)}
RAG was proposed to ground LLMs in external knowledge to mitigate hallucination~\cite{lewis2020retrieval}. Dense retrievers showed strong results~\cite{karpukhin2020dense, izacard2022few, gao2023retrieval}, but struggled with unstructured or redundant data~\cite{izacard2020leveraging, petroni2020kilt}. KG-RAG leveraged structured KGs~\cite{peng2024graph} to enable more precise evidence retrieval. GNN-RAG~\cite{mavromatis2024gnn} propagated information via graph neural networks, encoding queries and triples into a shared embedding space. SubgraphRAG~\cite{li2024simple} improved subgraph retrieval quality through lightweight structural scoring. Both approaches, however, operated in a single, monolithic embedding space and did not model relational structure at the level of individual reasoning hops. We instead focus on query–graph representations that explicitly decompose reasoning by hops.
 
\subsection{Multi-Head Embeddings}
Multi-head attention (MHA) variants improved embeddings across modalities, e.g., visual semantic alignment via self-attention~\cite{park2020mhsan}, redundancy reduction~\cite{bhojanapalli2021leveraging, bian2021attention}, and low-resource multilingual tasks~\cite{vashisht2025mage}. Recent interpretability work has shown that attention heads in LLMs tend to specialize in distinct semantic roles, exhibiting consistent patterns across inputs and architectures~\cite{zheng2025attentionheads, basile2025headpursuit}. In RAG, MRAG~\cite{besta2024multi} leveraged multiple attention heads as parallel retrieval channels to capture diverse semantic aspects of a query, working well for unstructured textual corpora. However, MRAG was designed around independent, non-aligned head channels, where query representations and context embeddings need to be projected into consistent semantic subspaces across reasoning hops. We address this gap by symmetrically decomposing both queries and KG triples into shared head-specific subspaces.

\subsection{Multi-Hop QA}
Prior methodologies generally bifurcated into enhancing LLM reasoning via decomposition or prompting~\cite{perez2020unsupervised, fu2021decomposing, wei2022chain}, and improving retrieval via iterative or adaptive methods~\cite{trivedi2022interleaving, asai2024self}. LLM-prompted methods such as StructGPT~\cite{jiang2023structgpt} and ToG~\cite{sun2023think} enabled step-wise KG traversal but required multiple LLM calls and incurred high latency. EtD~\cite{liu2024explore} reduced cost with a lightweight GNN retriever, but operated within a single-view embedding space. Path-based methods such as RoG~\cite{luo2023reasoning} planned relation paths with full supervision, treating retrieval and planning as disjoint stages. RoE~\cite{han2026roe} unified exploration and generation through Reinforcement Learning (RL) but incurred high trajectory training costs. Deep GraphRAG~\cite{li2026deepgraphrag} introduced hierarchical global-local retrieval via RL, yet required complex hierarchy maintenance. Across these methods, hop-level representation alignment between the query and the graph remains largely unaddressed. We approach this with an encoder-only Transformer trained under a weakly supervised objective, where multi-head attention is organized to reflect the hop structure of multi-hop reasoning.

\section{The ParallaxRAG Framework}

Multi-hop reasoning over knowledge graphs requires connecting multiple relational steps in a consistent manner. 
However, most KG-RAG systems collapse this process into a single embedding space, which obscures step-wise structure. 
ParallaxRAG builds on the intuition that each reasoning hop should be represented from a distinct perspective. 
By decomposing both the query and the graph into multiple, aligned views, and regulating their interaction through a balanced exploration–exploitation process, 
ParallaxRAG retrieves compact, interpretable subgraphs that better support multi-hop reasoning.

\subsection{Symmetric Decoupling of Query and Graph Views}

In standard KG-RAG, a complex query is encoded as a single vector $\mathbf{q}\in\mathbb{R}^d$, conflating semantics required for different reasoning stages. 
We instead leverage the internal multi-head structure of a Transformer to obtain $H$ specialized query views:
\begin{equation}\label{eq:query-views}
Q^{\text{views}}=\{\mathbf{q}_k\in\mathbb{R}^{d_h}\}_{k=1}^H.
\end{equation}
Concretely, we extract $\mathbf{q}_k$ from the \texttt{[CLS]} representation of the final Transformer layer: the $H$ attention heads naturally partition the hidden state into $H$ complementary subspaces of dimension $d_h = d/H$, each capturing a distinct semantic facet of the query. A shared linear projection $W_{\mathrm{proj}}$ maps each head slice to the same $d_h$-dimensional space and is applied symmetrically to entity and relation texts, yielding $\mathbf{e}_k, \mathbf{r}_k \in \mathbb{R}^{d_h}$.
This design aligns the query and the graph across multiple semantic dimensions, allowing each head to focus on a particular aspect such as entity grounding or relational chaining.
Unlike MRAG~\cite{besta2024multi}, which primarily leverages multiple heads as parallel embedding channels for multi-aspect retrieval, our symmetric decomposition explicitly enforces alignment between query and graphs within each head. This design encourages each head to specialize in a coherent semantic role shared by both the query and the graph, facilitating more structured hop-wise reasoning.

\begin{table*}[t]
\centering
\setlength{\tabcolsep}{3pt} 
\caption{Main retrieval recall evaluation results for different models on WebQSP and CWQ datasets.}
\label{tab:main_results}
\resizebox{\textwidth}{!}{%
    \begin{tabular}{l cc ccc cc ccc cc ccc}
    \toprule
    \multirow{4}{*}{Model} & \multicolumn{5}{c}{Shortest Path Triple Recall} & \multicolumn{5}{c}{GPT-4o Triple Recall} & \multicolumn{5}{c}{Answer Entity Recall} \\
    \cmidrule(lr){2-6} \cmidrule(lr){7-11} \cmidrule(lr){12-16}
    & \multicolumn{2}{c}{WebQSP} & \multicolumn{3}{c}{CWQ} & \multicolumn{2}{c}{WebQSP} & \multicolumn{3}{c}{CWQ} & \multicolumn{2}{c}{WebQSP} & \multicolumn{3}{c}{CWQ} \\
    \cmidrule(lr){2-3} \cmidrule(lr){4-6} \cmidrule(lr){7-8} \cmidrule(lr){9-11} \cmidrule(lr){12-13} \cmidrule(lr){14-16}
    & 1 & 2 & 1 & 2 & $\geq 3$ & 1 & 2 & 1 & 2 & $\geq 3$ & 1 & 2 & 1 & 2 & $\geq 3$ \\
    & (65.8\%) & (34.2\%) & (28.0\%) & (65.9\%) & (6.1\%) & (65.8\%) & (34.2\%) & (28.0\%) & (65.9\%) & (6.1\%) & (65.8\%) & (34.2\%) & (28.0\%) & (65.9\%) & (6.1\%) \\
    \midrule
    cosine similarity & 0.874 & 0.405 & 0.629 & 0.442 & 0.333 & 0.847 & 0.483 & 0.629 & 0.511 & 0.464 & 0.943 & 0.253 & 0.903 & 0.472 & 0.289 \\
    SR+NSM w E2E & 0.565 & 0.324 & - & - & - & 0.580 & 0.376 & - & - & - & 0.916 & 0.301 & - & - & - \\
    Retrieve-Rewrite-Answer & 0.064 & 0.046 & - & - & - & 0.062 & 0.061 & - & - & - & 0.745 & 0.729 & - & - & - \\
    RoG & 0.869 & 0.415 & 0.766 & 0.597 & 0.253 & 0.446 & 0.271 & 0.347 & 0.293 & 0.122 & 0.874 & 0.677 & 0.920 & 0.827 & 0.628 \\
    G-Retriever & 0.335 & 0.216 & 0.134 & 0.205 & 0.168 & 0.345 & 0.284 & 0.159 & 0.240 & 0.226 & 0.596 & 0.446 & 0.377 & 0.384 & 0.269 \\
    GNN-RAG & 0.532 & 0.502 & 0.515 & 0.498 & 0.446 & 0.384 & 0.445 & 0.328 & 0.408 & 0.418 & 0.810 & 0.831 & 0.853 & 0.841 & \textbf{0.787} \\
    SubgraphRAG & 0.954 & 0.720 & 0.845 & \textbf{0.826} & \textbf{0.609} & 0.895 & 0.768 & 0.787 & 0.810 & 0.725 & 0.979 & 0.844 & 0.937 & 0.918 & 0.683 \\ \midrule
    ParallaxRAG (Decoupled + Gated) & 0.963 & 0.756 & 0.891 & 0.809 & 0.568 & 0.916 & 0.820 & 0.829 & 0.842 & 0.759 & 0.976 & 0.884 & 0.958 & 0.928 & 0.753 \\
    ParallaxRAG & \textbf{0.966} & \textbf{0.761} & \textbf{0.916} & 0.818 & 0.578 & \textbf{0.923} & \textbf{0.825} & \textbf{0.847} & \textbf{0.845} & \textbf{0.760} & \textbf{0.986} & \textbf{0.899} & \textbf{0.962} & \textbf{0.935} & 0.771 \\
    \bottomrule
    \end{tabular}%
}

\end{table*}

\subsection{Balancing Exploration and Exploitation}

Creating multiple semantic views introduces flexibility but also tension: the model must encourage heads to explore distinct reasoning cues, yet prevent them from drifting toward irrelevant or redundant information. 
ParallaxRAG balances them through \emph{exploration} for head diversity and \emph{exploitation} for query relevance.

\paragraph{Exploration: Pairwise Similarity Regulation.}
To encode structural context, we adopt Distance Encoding (DE) (details in~\ref{app:psr}), which propagates signals from the topic entity via $L_f$ forward and $L_r$ reverse message-passing layers, yielding per-head structural features $\delta_k$.
At each DE layer~$\ell$, head $k$ integrates neighborhood context via:
\begin{equation}\label{eq:de-mp}
\tilde{\mathbf{F}}_k^{(\ell+1)}(v_i)=\!\sum_{v_j\in\mathcal{N}(v_i)}\!\frac{1}{|\mathcal{N}(v_i)|}\mathbf{F}_k^{(\ell)}(v_j)W^{(\ell)}.
\end{equation}
While this process spreads information across entities, multiple heads could converge to similar activation patterns.  
To maintain diversity, PSR measures the overlap among heads via their activation summaries:
\begin{equation}\label{eq:psr}
\mathbf{s}_k^{(\ell)}=\frac{\tilde{\mathbf{F}}_k^{(\ell+1)}\mathbf{1}_d}{\|\tilde{\mathbf{F}}_k^{(\ell+1)}\mathbf{1}_d\|_2},\qquad
r_k^{(\ell)}=\sum_{j\neq k}\langle\mathbf{s}_k^{(\ell)},\mathbf{s}_j^{(\ell)}\rangle.
\end{equation}
Each head’s update is then adaptively scaled by a diversity coefficient:
\begin{equation}\label{eq:psr-update}
\mathbf{F}_k^{(\ell+1)}=e^{-\beta r_k^{(\ell)}}\tilde{\mathbf{F}}_k^{(\ell+1)},
\end{equation}
where $\beta$ controls the penalty strength. This keeps the retrieval process exploratory, where heads remain distinct, each probing a different facet of the graph.

\paragraph{Head-Specific Triple Scoring.}
Under each head $k$, a candidate triple $\tau = (h, r, t)$ is scored by a 
shared MLP over the concatenation of the query view, augmented entity, and
relation representations, where $\tilde{\mathbf{e}}_k = [\mathbf{e}_k;\,\delta_k]$
denotes the entity embedding concatenated with its DE structural features, and
superscripts $h$ and $t$ refer to the head and tail entity of $\tau$ respectively:
\begin{equation}\label{eq:triple-score}
z_k(\tau) = \mathrm{MLP}\!\bigl(
[\mathbf{q}_k;\,\tilde{\mathbf{e}}_k^{h};\,
\mathbf{r}_k;\,\tilde{\mathbf{e}}_k^{t}]\bigr).
\end{equation}
Stacking scores across all triples and heads yields $Z \in \mathbb{R}^{|\mathcal{E}| \times H}$, 
which the query-aware gate then aggregates into a final ranking.

\paragraph{Exploitation: Query-Aware Gating with Weak Supervision.}
To focus retrieval on the most relevant heads for a given question, a lightweight gate maps the global query embedding $\mathbf{q}$ to head-importance weights
\begin{equation}\label{eq:gate}
\boldsymbol{\alpha}=\text{softmax}(W_g\mathbf{q}) \in \mathbb{R}^H.
\end{equation}
Each head’s shared MLP scores candidate triples via Eq.~\eqref{eq:triple-score}, yielding $Z\!\in\!\mathbb{R}^{|\mathcal{E}|\times H}$; gated aggregation produces
\begin{equation}\label{eq:gated-agg}
\mathbf{s}=Z\boldsymbol{\alpha},\qquad P_{\text{pred}}=\text{softmax}(\mathbf{s}).
\end{equation}
For each question, we extract the shortest paths linking its topic and answer entities in the KG. 
Triples on these paths ($\mathcal{T}_{\text{sp}}$) form the weak supervision signal, defining a normalized target distribution:
\begin{equation}\label{eq:p-true}
P_{\text{true}}(\tau)=
\frac{\mathbb{I}[\tau\in\mathcal{T}_{\text{sp}}]\,w_\tau}
{\sum_{\tau'\in\mathcal{E}}\mathbb{I}[\tau'\in\mathcal{T}_{\text{sp}}]\,w_{\tau'}}.
\end{equation}
The retriever is trained end-to-end by minimizing 
\begin{equation}\label{eq:loss}
\mathcal{L}=\text{KL}(P_{\text{true}}\|P_{\text{pred}}),
\end{equation}
which guides the gate and heads to emphasize triples composing the shortest topic–answer reasoning chains.

\subsection{Grounded Reasoning with LLMs}

The top-$k$ retrieved triples are linearized and concatenated with the question, together with a one-shot demonstration, to construct an evidence-grounded prompt for the LLM. 
This prompt formulation provides explicit relational context for each reasoning step and ensures that the model’s generation remains aligned with hop-specific evidence derived from the multi-view retriever. Details shows in Appendix \ref{app:prompt}.

\section{Experiments}
\label{sec:experiments}

We design a series of experiments to examine whether multi-view decoupling leads to more reliable and interpretable multi-hop reasoning. 
Specifically, we investigate three key questions:
\textbf{RQ1:} Does ParallaxRAG retrieve cleaner and more relevant subgraphs efficiently?
\textbf{RQ2:} Do multi-head views specialize in different reasoning stages and improve retrieval quality?
\textbf{RQ3:} Can ParallaxRAG enhance answer grounding and reduce hallucination in end-to-end KGQA?
Sections~\ref{sec:retrieval}, \ref{sec:e2e} and \ref{sec:bioasq} address RQ1, Section~\ref{sec:heads} analyzes RQ2, and Section~\ref{sec:e2e} further studies RQ3.

\subsection{Experiment Setup}
\label{sec:setup}
\paragraph{Datasets.}
We use WebQSP~\cite{yih2016web} and CWQ~\cite{talmor2018web} as 
benchmarks and further test cross-domain generalization on 
BioASQ~\cite{tsatsaronis2015overview}, which requires biomedical 
factual reasoning.
\paragraph{Baselines.}
\subparagraph{Retrieval Baselines.}
We compare against representative retrieval paradigms, structure-free (Cosine Similarity~\cite{li2024simple}), path-based (SR+NSM~\cite{zhang2022subgraph}, Retrieve-Rewrite-Answer~\cite{wu2023retrieve}, RoG~\cite{luo2023reasoning}), hybrid (G-Retriever~\cite{he2024g}), and GNN-based (GNN-RAG~\cite{mavromatis2024gnn}), spanning a broad spectrum from flat vector matching to graph-structured reasoning.

\subparagraph{End-to-End KGQA Baselines.}
We further evaluate against state-of-the-art KGQA models—UniKGQA~\cite{jiang2022unikgqa}, 
KD-CoT~\cite{zhao2024kg}, 
StructGPT~\cite{jiang2023structgpt}, 
ToG~\cite{sun2023think}, 
EtD~\cite{liu2024explore}, 
SubgraphRAG\cite{zhang2022subgraph}, 
REL-RAG\cite{yao2025learningefficientgeneralizablegraph} 
and 
GraphRAG-FI\cite{guo2025graphragfi}
Following~\cite{zhang2022subgraph}, we adopt the RoG-Sep variant to avoid label leakage in RoG-Joint training.

\paragraph{Implementation Details.}
All retrievers use the \texttt{BGE-M3}\footnote{\url{https://huggingface.co/BAAI/bge-m3}} encoder for consistency. 
Baselines are reproduced from official implementations. 
Key hyperparameters and training details are listed in Appendix~\ref{app:exp_details}. 
Results using alternative backbones are shown in Appendix~\ref{app:backbone}.

\paragraph{Metrics.}
We report both retrieval and end-to-end performance.  
\textit{Retrieval:} (i) shortest-path triple recall, 
(ii) GPT-4o–verified triple recall (validating the correctness of recalled triples), and 
(iii) answer-entity recall, averaged per query and broken down by hop length, we report two ParallaxRAG configurations, one with the decoupled-and-gated architecture alone and one with additional PSR, to disentangle the structural contribution from regularization effects.
\textit{Efficiency:} wall-clock inference time on a 48GB RTX 6000 Ada GPU (excluding KG I/O).  
\textit{End-to-end QA:} Macro-F1 and Hit scores on WebQSP/CWQ, plus domain metrics on BioASQ.

\subsection{Retrieval Performance (\textbf{RQ1})}
\label{sec:retrieval}

Notably, even without PSR, the decoupled-and-gated ParallaxRAG configuration already outperforms almost all baseline retrievers across both datasets, especially on Answer Entity Recall, indicating that the core multi-view architecture itself provides a strong inductive bias for multi-hop retrieval.
Pairwise Similarity Regulation (PSR) further strengthens this effect, particularly on longer-hop queries, by preventing head collapse and improving robustness under combinatorial expansion.

In terms of efficiency, ParallaxRAG remains computationally lightweight, requiring 40 seconds on WebQSP and 84 seconds on CWQ (excluding KG I/O), compared to 948 and 2327 seconds for RoG, 672 and 1530 seconds for G-Retriever.

\begin{table}[htbp]
\centering
\scriptsize
\setlength{\tabcolsep}{3pt}
\caption{Question-answering performance on WebQSP and CWQ. 
The Hallucination (Hallu.) score is evaluated on a subset that excludes samples where the answer entity is not in the KG, following \cite{li2024simple}. 
Our generaters use the top 100 retrieved triples by default; results for 200 and 500 (indicated in parentheses) are also shown. 
Best results are in \textbf{bold}. 
Results with ($\leftrightarrow$) evaluate retriever generalizability, where the retriever is trained on one dataset and applied to the other.}
\label{tab:qa_performance_webqsp_cwq}
\textbf{\setlength{\tabcolsep}{1pt}}
\resizebox{\columnwidth}{!}{%
\begin{tabular}{lcccccc}
\toprule
 & \multicolumn{3}{c}{WebQSP} & \multicolumn{3}{c}{CWQ} \\
 \cmidrule(lr){2-4} \cmidrule(lr){5-7}
\textbf{Model} & \textbf{Macro-F1} & \textbf{Hit} & \textbf{Hallu.} & \textbf{Macro-F1} & \textbf{Hit} & \textbf{Hallu.} \\
\midrule
\multicolumn{7}{l}{\textit{(A) Neural Methods}} \\
UniKGQA & 72.2 & - & - & 49.0 & - & - \\
SR+NSM w E2E & 64.1 & - & 64.44 & 46.3 & - & - \\

\midrule
\multicolumn{7}{l}{\textit{(B) Multi-turn LLM Reasoning Methods}} \\
KD-CoT & 52.5 & 68.6 & - & - & 55.7 & - \\
ToG (GPT-4) & - & 82.6 & - & - & 67.6 & - \\
StructGPT & - & 74.69 & - & - & - & - \\
Retrieve-Rewrite-Answer & - & 79.36 & - & - & - & - \\
RoG-Joint & 70.26 & 86.67 & 76.13 & 54.63 & 61.94 & 55.15 \\
RoG-Sep & 66.45 & 82.19 & 72.79 & 53.87 & 60.55 & 54.51 \\
RoG + GraphRAG-FI & 73.86 & 89.25 & - & 55.12 & 64.82 & - \\
\midrule
\multicolumn{7}{l}{\textit{(C) RAG-based Methods}} \\
G-Retriever & 53.41 & 73.46 & 67.97 & - & - & - \\
EtD & - & 82.5 & - & - & 62.0 & - \\
GNN-RAG & 71.3 & 85.7 & - & 59.4 & 66.8 & - \\
SubgraphRAG + GPT-4o  & 76.46 & 89.80 & 81.85 & 59.08 & 66.69 & 66.57 \\
REL-RAG + GPT-4o-mini & 78.7 & 92.5 & - & 58.6 & 68.3 & - \\
\midrule

\multicolumn{7}{l}{\textit{(D) Ours Methods}} \\
ParallaxRAG + Llama3.1-8B & 71.73 & 86.85 & \textbf{83.64} & 48.33 & 58.41 & 66.12 \\
ParallaxRAG + Qwen3-30B & 75.24 & 91.60 & 75.45 & 59.25 & 65.30 & 57.18 \\
ParallaxRAG + Qwen3-30B (200) & 76.07 & 92.48 & 76.23 & 59.31 & 66.21 & 58.06 \\
ParallaxRAG + Qwen3-30B (500) & 76.11 & 93.26 & 77.86 & 59.18 & 64.52 & 60.07 \\
ParallaxRAG + GPT-4o (200) & \textbf{78.80} & \textbf{93.53} & 82.94 & \textbf{62.31} & \textbf{70.74} & \textbf{66.69} \\
\midrule
\multicolumn{7}{l}{\textit{(E) Cross-dataset Generalization ($\leftrightarrow$)}} \\
SubgraphRAG + Llama3.1-8B ($\leftrightarrow$) & 66.42 & 83.42 & 80.09 & 37.96 & 48.57 & 56.78 \\
ParallaxRAG + Llama3.1-8B ($\leftrightarrow$) & \textbf{69.22} & \textbf{89.51} & 82.64 & \textbf{45.28} & \textbf{54.82} & 60.04 \\
\bottomrule
\end{tabular}
} 
\end{table}

\begin{figure*}[!t]
  \centering
  \begin{subfigure}{\textwidth}
    \centering
    \includegraphics[width=1\textwidth]{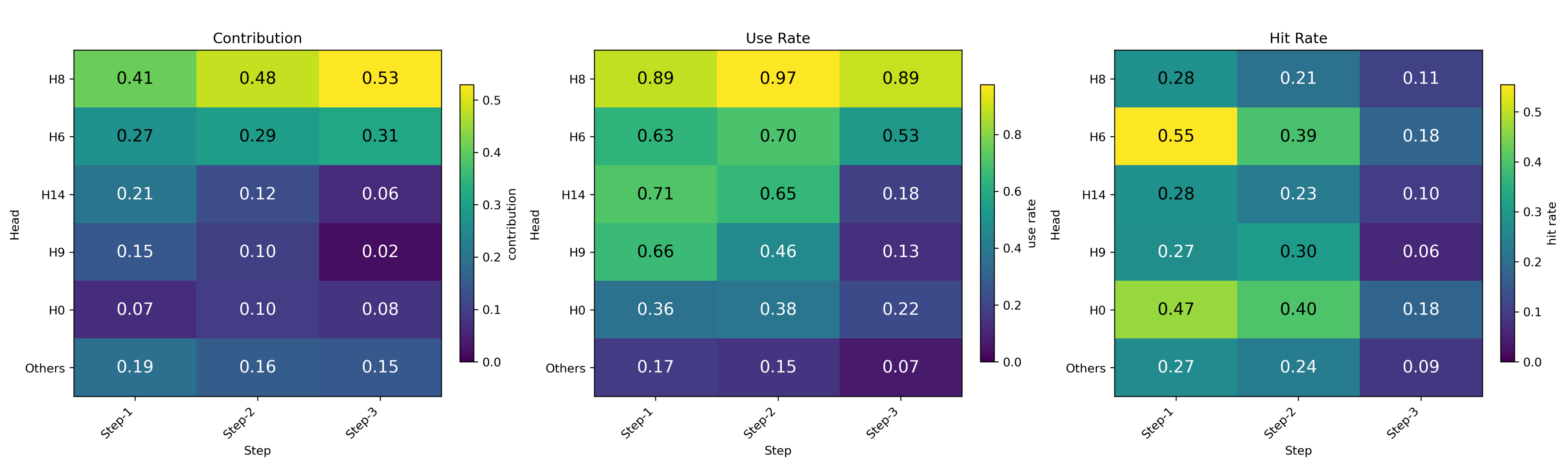}
    \caption{WebQSP: Early-Step Specialization for Short-Chain Reasoning}
    \label{fig:heat_1}
  \end{subfigure}

  \begin{subfigure}{\textwidth}
    \centering
    \includegraphics[width=1\textwidth]{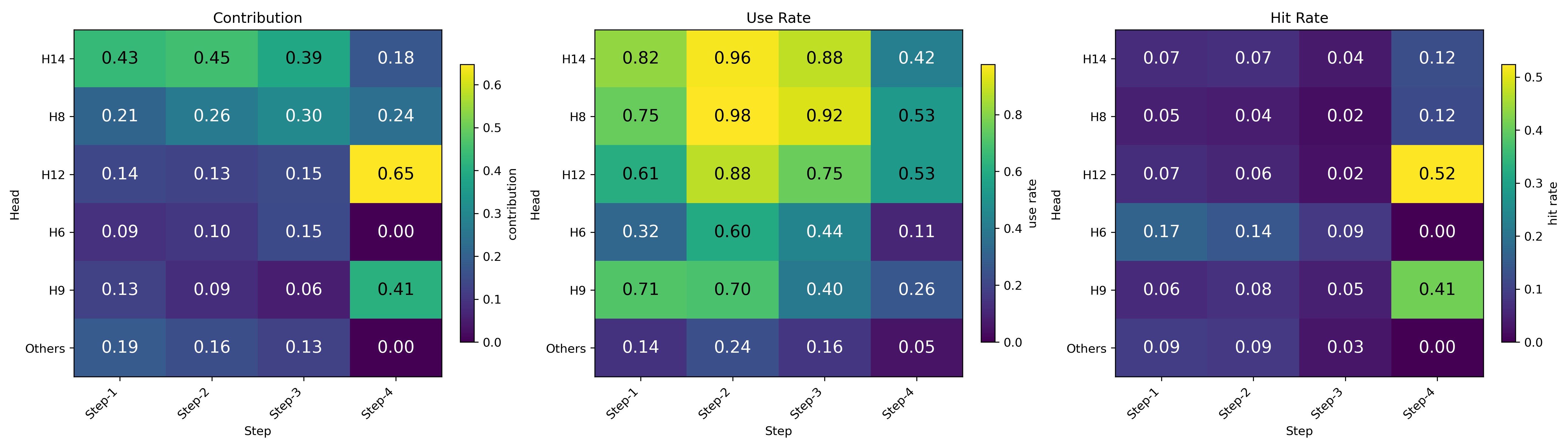}
    \caption{CWQ: Dynamic Head Specialization Switching for Long-Chain Reasoning}
    \label{fig:heat_2}
  \end{subfigure}

  \caption{Visualization of attention head specialization, revealing a task-adaptive division of labor. The heatmaps display three metrics (Contribution, Use Rate, Hit Rate) for the top 5 specialist heads, selected from our BGE-based model's 16 heads according to their overall contribution. The Others row aggregates the remaining 11 heads: for Contribution, this value is the sum of their contributions, while for Use Rate and Hit Rate, it is their average, see more explanation in (Appendix~\ref{app:head}). We define \emph{steps} as the BFS expansion depth along reasoning paths, rather than the conventional hop (shortest-path length), in order to capture head behaviors at each layer of candidate exploration. \textbf{(a)} On WebQSP (short chains). \textbf{(b)} On CWQ (long chains).}
  \label{fig:heat_all}
\end{figure*}

\begin{table}[h]
\centering
\caption{Answer Entity Recall under controlled disruption on CWQ, broken down by hop depth. Long-hop-associated heads are functionally non-substitutable.}
\label{tab:disruption}
\footnotesize
\resizebox{\linewidth}{!}{%
\begin{tabular}{lcccc}
\toprule
& \multicolumn{4}{c}{Answer Entity Recall} \\
\cmidrule(lr){2-5}
\textbf{Condition} & \textbf{1-hop} & \textbf{2-hop} & \textbf{$\geq$3-hop} & \textbf{Overall} \\
\midrule
ParallaxRAG Baseline          & 0.959 & 0.926 & 0.763 & 0.883 \\
Drop Long-Hop Heads     & 0.958 & 0.777 & 0.501 & 0.745 \\
Shuffle Heads          & 0.734 & 0.532 & 0.468 & 0.578 \\
All               & 0.734 & 0.446 & 0.421 & 0.534 \\
\bottomrule
\end{tabular}%
}
\end{table}

\begin{table*}[htbp]
\centering
\caption{Generalization on BioASQ Task~B, where the retriever is trained on CWQ and transferred without fine-tuning. We compare the performance of Qwen3-Plus, Single-View and ParallaxRAG (Multi-View), isolating the contribution of the retrieval design}
\label{tab:bioasq_generalization}
\resizebox{\textwidth}{!}{%
\begin{tabular}{lcccccccccc}
\toprule
\multirow{2}{*}{Model} & \multicolumn{4}{c}{Yes/No Questions} & \multicolumn{3}{c}{Factoid Questions} & \multicolumn{3}{c}{List Questions} \\
\cmidrule(lr){2-5} \cmidrule(lr){6-8} \cmidrule(lr){9-11}
 & Acc. & F1 Yes & F1 No & Macro F1 & Strict & Lenient & MRR & Prec. & Recall & F1 \\
\midrule
Deepseek-R1-8B & 0.9077 & 0.9312 & 0.8558 & 0.8935 & 0.2620 & 0.2733 & 0.2677 & 0.3982 & 0.3427 & 0.3517 \\
GPT4-Turbo & 0.9129 & 0.9357 & 0.8616 & 0.8986 & \textbf{0.5505} & 0.6515 & 0.6010 & 0.5788 & \textbf{0.4857} & \textbf{0.5051} \\
GPT-4o & 0.9140 & 0.9347 & 0.8670 & 0.9009 & 0.3462 & 0.3462 & 0.3462 & 0.5102 & 0.4025 & 0.4330 \\
Qwen3-Plus  & 0.9012 & 0.9255 & 0.8402 & 0.8829 & 0.3219 & 0.6213 & 0.4132 & 0.5682 & 0.3989 & 0.4563 \\
\midrule
Single-View + Qwen3-Plus    & 0.9112 & 0.9267 & 0.8374 & 0.8821 & 0.3773 & 0.7214 & 0.8462 & 0.6833 & 0.4280 & 0.4464 \\
ParallaxRAG + Qwen3-Plus & \textbf{0.9351} & \textbf{0.9524} & \textbf{0.8980} & \textbf{0.9252} & 0.4210 & \textbf{0.8222} & \textbf{0.8667} & \textbf{0.7116} & 0.4569 & 0.4737 \\
\bottomrule
\end{tabular}
} 
\end{table*}

\begin{table*}[!t]
\centering
\caption{
Ablation study on WebQSP and CWQ, analyzing the impact of ParallaxRAG's core multi-head architecture and its synergistic mechanisms. 
Values in parentheses ($\downarrow$) indicate the performance drop compared to the full model (Llama3.1-8B as generator).
}
\label{tab:ablation}
\footnotesize
\begin{tabularx}{\textwidth}{l >{\raggedright\arraybackslash}X c c c c}
\toprule
& & \multicolumn{2}{c}{WebQSP} & \multicolumn{2}{c}{CWQ} \\
\cmidrule(lr){3-4} \cmidrule(lr){5-6}
& Configuration & Macro-F1 & Hit & Macro-F1 & Hit \\
\midrule
& ParallaxRAG (Full Model)
& \textbf{71.73} & \textbf{86.85} & \textbf{48.33} & \textbf{58.41} \\
\midrule
\multicolumn{6}{l}{Multi-Head Architecture Validation} \\
(a) & Split Vector Baseline 
& 69.42 ($\downarrow$2.31) & 85.02 ($\downarrow$1.83) & 42.27 ($\downarrow$6.06) & 51.54 ($\downarrow$6.87) \\
(b) & Single Vector Baseline 
& 70.60 ($\downarrow$1.13) & 86.11 ($\downarrow$0.74) & 44.82 ($\downarrow$3.51) & 52.63 ($\downarrow$5.78) \\
\midrule
\multicolumn{6}{l}{Synergistic Mechanism Validation} \\
(c) & Without PSR 
& 69.73 ($\downarrow$2.00) & 85.47 ($\downarrow$1.38) & 47.12 ($\downarrow$1.21) & 56.32 ($\downarrow$2.09) \\
(d) & Without Query-Aware Gating 
& 56.25 ($\downarrow$15.48) & 74.69 ($\downarrow$12.16) & 29.14 ($\downarrow$19.19) & 38.26 ($\downarrow$20.15) \\
\bottomrule
\end{tabularx}
\end{table*}





\subsection{Attention Head Specialization \textbf{(RQ2)}}
\label{sec:heads}

Figure~\ref{fig:heat_all} reveals a clear division of labor among ParallaxRAG’s attention heads that adapts to query complexity. 
For short-hop questions in WebQSP, early heads dominate across 
reasoning steps. As the reasoning chain extends in CWQ, later heads 
become increasingly active at deeper reasoning steps. This dynamic 
shift forms a relay pattern, where distinct head groups take over 
successively as reasoning deepens.

We quantify this specialization by training a linear probe to predict reasoning depth from head activations, achieving 52.3\% accuracy compared to a 25\% random baseline, demonstrating that head activations encode distinct reasoning stages. 
To examine the functional role of specialist heads, we perform a Difference-in-Differences-in-Differences (DDD) analysis comparing performance drops after ablating specialist versus random heads across short and long queries:
\begin{equation*}
\footnotesize
\resizebox{0.98\linewidth}{!}{$
\text{DDD} = [(\Delta_{\text{Long}}^{\text{Specialist}} - \Delta_{\text{Short}}^{\text{Specialist}})]
- [(\Delta_{\text{Long}}^{\text{Random}} - \Delta_{\text{Short}}^{\text{Random}})]
$}
\end{equation*}
where $\Delta$ denotes the post-ablation performance difference. 
A significant negative DDD value of -0.0184 (95\% CI: [-0.0248, -0.0045], $p$=0.0055) provides functional evidence that specialist heads are non-substitutable, playing distinct roles in multi-hop reasoning.

To further confirm this, we conduct controlled disruption experiments on CWQ under three conditions: \textbf{Drop Long-Hop Heads} removes the long-hop-associated head groups (8,9,12) identified from Figure~\ref{fig:heat_all}; \textbf{Shuffle Heads} randomly permutes the head-to-query-view assignment; and \textbf{All} applies both simultaneously. As Table~\ref{tab:disruption} shows, dropping long-hop heads has negligible effect on 1-hop queries but substantially degrades $\geq$3-hop performance by 0.262 absolute. Shuffling head alignment impairs all hop depths, with degradation scaling with query complexity. These asymmetric patterns confirm that the identified head groups play coordinated, non-interchangeable roles in multi-hop reasoning over knowledge graphs.

\subsection{End-to-End KGQA with RAG (\textbf{RQ1} \& \textbf{RQ3})}
\label{sec:e2e}

We next examine whether the retrieval and head-specialization advantages of ParallaxRAG lead to improvements in downstream question answering. 
As shown in Table~\ref{tab:qa_performance_webqsp_cwq}, when paired with Llama3.1-8B and 100 retrieved triples, ParallaxRAG attains a Macro-F1 of 71.73 and Hit of 86.85 on WebQSP, and 48.33 Macro-F1 and 58.41 Hit on CWQ. Combined with GPT-4o and 200 triples, it establishes new state-of-the-art results, reaching 78.80 Macro-F1 and 93.53 Hit on WebQSP, and 62.31 Macro-F1 and 70.74 Hit on CWQ. The gain from a stronger generator is disproportionately larger on CWQ, suggesting that for complex multi-hop chains the generator's capacity to integrate multi-relational evidence becomes the binding constraint once retrieval quality is sufficient. Expanding the retrieval budget from 100 to 200 triples yields moderate gains on both datasets, while further expansion to 500 triples provides negligible improvement on WebQSP and slightly hurts CWQ, indicating that retrieval precision rather than coverage is the key driver of downstream accuracy.

Beyond overall accuracy, ParallaxRAG improves reliability by reducing hallucinated generations. 
On hallucination-controlled subsets~\citep{li2024simple}, hallucination scores decrease by 1.22\% on WebQSP and 3.23\% on CWQ, indicating that the retrieved subgraphs are not only sufficient but also cleaner and more grounded. 
Qualitative examples in Appendix~\ref{app:case} further illustrate this effect: for long-chain CWQ questions, ParallaxRAG retrieves fewer but more relevant triples, enabling coherent reasoning chains that accurately support the final answers. To assess reproducibility, we conducted five independent runs and observed stable performance (standard deviation below 1.4 Macro-F1 points). Improvements over strong baselines are statistically significant via paired bootstrap resampling ($p < 0.05$); full details are reported in Appendix~\ref{app:stats}.

\subsection{Domain Generalization (\textbf{RQ1)}}
\label{sec:bioasq}

To evaluate cross-domain robustness, we test ParallaxRAG on the biomedical BioASQ benchmark (Task~B), which covers Yes/No, Factoid, and List questions. The retriever is trained on CWQ and transferred without retraining. We use Qwen3-Plus as the generator and report official metrics for each question type.

As shown in Table~\ref{tab:bioasq_generalization}, ParallaxRAG achieves the best Yes/No results, with an accuracy of 0.9351 and a Macro-F1 of 0.9252, outperforming strong general-purpose LLMs. On Factoid questions, it attains a lenient accuracy of 0.8222 and an MRR of 0.8667, reflecting effective multi-view reasoning even under domain shift. For List questions, ParallaxRAG favors precision over recall, yielding a slightly lower F1 than GPT-4-Turbo, which suggests a bias toward concise, high-confidence retrieval.

We further compares single-view and multi-view configurations under the same generator, prompting format, and context budget, confirming that the gains stem from the multi-view retrieval design rather than from the choice of generator.

\subsection{Ablation Study}
First, to verify the role of the multi-head design, we compare two simplified baselines. 
The Split Vector variant divides a single query embedding into pseudo-heads without true head specialization, causing a Macro-F1 drop of 2.31 on WebQSP and 6.06 on CWQ. 
Similarly, the Single Vector variant using a flat embedding, shows performance drops of 1.13 and 3.51 points, confirming that learned multi-head representations are crucial for capturing distinct relational cues and mitigating representational collapse.

Next, we evaluate the synergistic mechanisms. 
Removing Pairwise Similarity Regularization (PSR) yields redundant head behavior and reduces Macro-F1 by 2.00 and 1.21 points, demonstrating PSR’s role in maintaining head diversity. 
Eliminating the query-aware gating and replacing it with simple averaging leads to severe performance degradation, with Hit rate drops of 12.16 and 20.15 points, underscoring its necessity for dynamic head weighting and precise retrieval.

Together, removing either component degrades performance substantially, with gating removal causing the sharpest drop (Hit $-$12.16/$-$20.15), indicating it is the more critical of the two.

\section{Conclusion}

We propose ParallaxRAG, a multi-view KG-RAG framework that explicitly leverages attention head specialization for multi-hop reasoning. By decoupling queries and knowledge graphs into aligned head-specific semantic spaces and combining diversity regularization with query-aware aggregation, ParallaxRAG retrieves accurate and clean subgraphs while remaining efficient. Experiments demonstrate state-of-the-art retrieval and QA performance on WebQSP and CWQ, reduced hallucination, and zero-shot generalization to the biomedical BioASQ benchmark. Beyond performance gains, our analysis provides functional evidence suggesting that attention heads tend to specialize in different stages of multi-hop reasoning and play coordinated, non-substitutable roles within the structured multi-view framework. These results suggest that modeling head-level semantic specialization is a practical and generalizable direction for retrieval-augmented multi-hop reasoning over KGs.

\section{Limitations and Future Work}

While ParallaxRAG demonstrates strong retrieval performance and effectively captures head-level specialization, we identify several limitations that suggest directions for future work.

First, our method relies on the quality of the underlying knowledge graph and embedding model; errors in these components may propagate to retrieval performance. In addition, the current implementation focuses on static knowledge graphs and does not address dynamic or continuously updated settings. We leave extensions to noisier and larger-scale knowledge sources as future work.

Second, our training relies on shortest-path triples as weak supervision signals, following established practice in prior work~\cite{luo2023reasoning}. While shortest paths are a practical and scalable source of supervision, they may not always represent the optimal reasoning route, and this choice may introduce a bias toward shortest-path evidence. That said, our primary evaluation metric, Answer Entity Recall, requires only that the gold answer entity appear in the retrieved subgraph regardless of the specific path taken, which partially mitigates this concern. Developing retrieval supervision that incorporates diverse, multi-path annotations remains an important direction for future work.

Third, as illustrated by our case study, the benefits of multi-view retrieval are most pronounced for complex multi-hop queries, while they may be less critical for shallow queries (e.g., 1-hop transitive relations). In such cases, retrieving a richer set of evidence across multiple views can introduce contextual signals that are not strictly required for answering the question. This may increase the burden on downstream LLMs to reconcile partially redundant information, occasionally resulting in conservative predictions. This observation aligns with the observation proposed by \cite{guo2025graphragfi}. A direction for future work is to develop adaptive retrieval strategies that dynamically modulate retrieval breadth based on estimated query complexity.

\section*{Acknowledgments}

This work was supported by the National Natural Science Foundation of China (Grant NO. 62576292), the Zhejiang Province Leading Geese Plan (2025C02025), the Science and Technology Program of Huzhou (Grant NOs. 2023GZ42 and 2024GZ09), and in part by the Yangtze Delta Region Institute (Huzhou) Guidance Fund of University of Electronic Science and Technology of China (Grant NO. U03210054).





\bibliography{custom}

\appendix
\section{Implementation Details of the ParallaxRAG Framework}
\subsection{Detailed Derivation of Pairwise Similarity Regulation (PSR)}
\label{app:psr}

The Pairwise Similarity Regulation (PSR) mechanism is integrated into each layer of the Distance Encoding (DE) to foster representational diversity.

\paragraph{1. DE Propagation.}
Given the initial one-hot topic entity features $\mathbf{X}_0 \in \mathbb{R}^{N \times 2}$, the model performs $L_f$ forward and $L_r$ reverse propagation layers. Each layer $\ell$ computes a preliminary update $\tilde{\mathbf{F}}_{k}^{(\ell+1)}$ for each head $k$ via message passing ($\text{MP}$):
\begin{equation}\label{eq:app-de-mp}
\begin{aligned}
\tilde{\mathbf{F}}_{k}^{(\ell+1)} &= \text{MP}(\mathbf{E}, \mathbf{F}_{k}^{(\ell)}) \\
&\quad (\text{forward pass, using edge index } \mathbf{E})
\end{aligned}
\end{equation}
The reverse pass is analogous, using the transposed edge index $\mathbf{E}^T$.
\paragraph{2. PSR Computation and Application.}
After each propagation step, the preliminary updates $\tilde{\mathbf{H}}$ are modulated by PSR. The process begins by computing a node intensity vector $\mathbf{s}_k^{(\ell)}$ for each head to capture its activation distribution:
\begin{equation}\label{eq:app-psr-s}
\mathbf{s}_k^{(\ell)} = \text{L2Norm}\left(\sum_{d} \tilde{\mathbf{F}}_{k}^{(\ell)}[:, d]\right)
\end{equation}
These vectors are then used to derive a redundancy score $r_k^{(\ell)}$ from pairwise similarities (cf.\ Eq.~\eqref{eq:psr}), which in turn defines the regulation coefficient $\alpha_k^{(\ell)}$:
\begin{equation}\label{eq:app-psr-r}
\begin{aligned}
r_k^{(\ell)} &= \sum_{j \neq k} \langle \mathbf{s}_i^{(\ell)}, \mathbf{s}_j^{(\ell)} \rangle \\
\alpha_k^{(\ell)} &= \exp(-\beta \cdot r_k^{(\ell)})
\end{aligned}
\end{equation}
Finally, this coefficient performs the final update by scaling the preliminary update to produce the final layer output:
\begin{equation}\label{eq:app-psr-final}
\mathbf{F}_{k}^{(\ell+1)} = \alpha_k^{(\ell)} \cdot \tilde{\mathbf{F}}_{k}^{(\ell+1)}
\end{equation}

\paragraph{3. Final Representation.}
The final structural representations $\mathcal{F}_k$ for each head are formed by collecting the outputs $\{\mathbf{F}_{k}^{(\ell)}\}$ from all layers, which are then used for triple scoring.

\subsection{Weighted Listwise Training Objective}
\label{app:loss}

To handle the severe class imbalance in retrieval, we convert the binary weak supervision signal $\mathbf{y} \in \{0,1\}^{|\mathcal{E}|}$ into a weighted target distribution for our listwise objective.

\paragraph{1. Positive Reweighting.}
The binary vector $\mathbf{y}$ is first normalized. Then, a weight factor $w_\tau$ is applied to construct the final weighted distribution. In our implementation, we use $w_\tau=10$ for positive triples ($y_\tau=1$) and $w_\tau=1$ otherwise. This is conceptually similar to the $\alpha$-balancing in Focal Loss~\cite{lin2017focal}.
\begin{equation}\label{eq:app-reweight}
y_\tau^{\text{norm}} = \frac{y_\tau}{\sum_{\tau'} y_{\tau'}}, \quad y_\tau^{\text{weighted}} = \frac{y_\tau \cdot w_\tau}{\sum_{\tau'} (y_{\tau'} \cdot w_{\tau'})}
\end{equation}

\paragraph{2. Final Loss Formulation.}
The model is trained by minimizing the weighted listwise cross-entropy (equivalent to Eq.~\eqref{eq:loss}) between the predicted distribution $P_{\text{pred}}$ and the target $y^{\text{weighted}}$:
\begin{equation}\label{eq:app-loss}
\mathcal{L}_{\text{listwise}} = -\sum_{\tau \in \mathcal{E}} y_\tau^{\text{weighted}} \,\log \big( P_{\text{pred}}(\tau) + \epsilon \big)
\end{equation}

\section{Additional Experiment Setting}

\label{app:exp_details}



\subsection{Implementation Details for ParallaxRAG}
Our model is implemented in PyTorch and PyTorch Geometric. The ParallaxRAG retriever, which consists of a \texttt{BGE-M3} text encoder, a PSR-enhanced DE module, and a scoring MLP, is trained end-to-end for up to 100 epochs with an early stopping patience of 20. The forward and reverse propagation round is set to 2 rounds, with PSR strength of $0.5$. We use the AdamW optimizer with a half-cycle cosine annealing with warmup learning rate, during a warmup phase for the first few epochs, the learning rate linearly increases from 0 to a peak value of $1\times10^{-3}$. Following the warmup, the learning rate is smoothly decayed along a cosine curve to a minimum value of $1\times10^{-5}$. to enable finer parameter adjustments in the later stages of training. An effective batch size of 2 (via gradient accumulation), and employ a weighted listwise ranking loss.

For the end-to-end KGQA task, the top-100 triples retrieved are linearized into a natural text format and prepended to the question as context for a LLM generator. We generate final answers using nucleus sampling with $p=0.95$ and a temperature of $0.7$.

\subsection{GPT-4o Triple Recall Verification Protocol}
\label{app:gpt_protocol}

In the main retrieval evaluation, GPT-4o is used solely as a structured extraction tool to identify which retrieved triples are necessary for answering a given question (GPT-4o-verified Triple Recall, $\mathcal{R}_{\mathrm{gpt}}$).
It is \emph{not} used as an answer generator or for any open-ended evaluation.
The exact prompt used for supporting-triple extraction is as follows.

\begin{tcolorbox}[
  title= Prompt format used for GPT-4o triple-recall verification.,
  fonttitle=\bfseries,
  colback=white,
  colframe=black!60,
  colbacktitle=black!70,
  coltitle=white,
  boxrule=0.5pt,
  arc=3pt
]
\ttfamily\footnotesize
\textbf{System Prompt:} Based on the triplets retrieved from a knowledge graph, please select the relevant triplets necessary for answering the question. Return the selected triplets as a list, each prefixed with ``evidence:''.

\medskip
\textbf{User:} \textit{[Retrieved triples]}\\
\textit{[Question from the dataset]}
\end{tcolorbox}

After extraction, triple recall is computed deterministically via scripted string matching against annotated ground truth, so the final metric does not depend on GPT-4o's generative tendencies.
To assess extraction stability, we repeated the extraction with three different models (GPT-4o, Qwen3-30B, GLM-4.7) and observed negligible variance (std $\leq$ 1.5\%).
We also manually inspected a random 10\% subset and found high consistency between automated extraction and human judgment.

\subsection{Prompt for Answer Generation}
\label{app:prompt}
The detailed prompt template used in our experiments for answer generation is as follows.

\begin{tcolorbox}[
  float,
  breakable,
  title=Prompt format used for QA.,
  fonttitle=\bfseries,
  colback=white,
  colframe=black!60,
  colbacktitle=black!70,
  coltitle=white,
  boxrule=0.5pt,
  arc=3pt,
  label=box:qa_prompt
]
\ttfamily\footnotesize

\textbf{System Prompt}\\
Based on the triplets from a knowledge graph, please answer the given question.
Please keep the answers as simple as possible and return all the possible answers
as a list, each with a prefix ``ans:''.

\tcbline  

\textbf{In-Context Learning (Few-shot) Examples}\\[4pt]
Triplets:\\
(Lou Seal, sports.mascot.team, San Francisco Giants)\\
(San Francisco Giants, sports.sports team.championships, 2012 World Series)\\
(San Francisco Giants, sports.sports championship event.champion, 2014 World Series)\\
(San Francisco Giants, time.participant.event, 2014 Major League Baseball season)\\
...\\[4pt]
Question:\\
What year did the team with mascot named Lou Seal win the World Series?\\[4pt]
To find the year ...... Therefore, the formatted answers are:\\
ans: 2014 (2014 World Series)\\
ans: 2012 (2012 World Series)\\
ans: 2010 (2010 World Series)

\tcbline  

\textbf{User Prompt}\\[4pt]
Triplets:\\
$(e_a, r_{ab}, e_b)$,\\
$(e_c, r_{cd}, e_d)$,\\
...\\[4pt]
Question:\\
What \ldots?

\end{tcolorbox}

\subsection{Hyperparameter Sensitivity Analysis}
\label{app:sensitivity}
In this section, we analyze the sensitivity of our model to the key hyperparameter $\beta$, which controls the strength of the Pairwise Similarity Regulation (PSR). We argue that a challenging zero-shot transfer task is the most effective setting to demonstrate PSR's true diversity impact, as it enables the model to maximally leverage its learned head specialization. Therefore, we evaluate a model trained on WebQSP on the CWQ test set, varying $\beta$ within the range of $\{ 0.2, 0.5, 2.0\}$, with $\beta=0$ serving as the baseline without diversity regulation.

\begin{figure}[htbp] 
    \centering
    \includegraphics[width=\columnwidth]{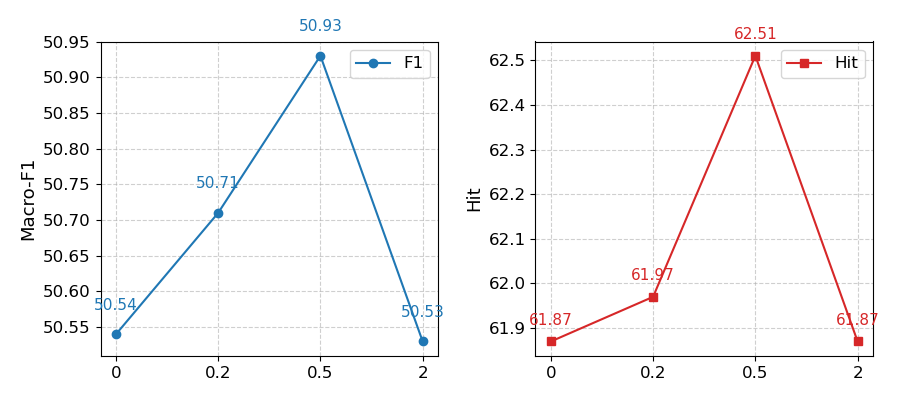}
    \caption{Sensitivity analysis of the PSR strength parameter $\beta$ on the generalization task, where the retriever was trained on WebQSP and tested on CWQ-sub, using Llama-3.1-8B as the generator.}
    \label{fig:sensitivity_beta}
\end{figure}

As shown in Figure~\ref{fig:sensitivity_beta}, the results under this stringent evaluation setting are revealing. A moderate PSR strength of $\beta=0.5$ achieves the peak performance, boosting the Macro-F1 from a baseline of 50.54 to 50.93 and the Hit rate from 61.87 to 62.51. This performance gain suggests that the diverse representations fostered by PSR are indeed learning more generalizable reasoning mechanisms. The degradation at a high penalty ($\beta=2.0$) indicates the limit of this effect, where excessive suppression can hinder signal propagation. These findings validate our choice of $\beta=0.5$ and confirm that the generalization setting is an effective testbed for evaluating the impact of PSR.

\section{Attention Head Analysis Details}
\label{app:head}

This section provides the detailed methodology for the attention head analysis presented in Section~\ref{sec:heads}. We formally define the concepts of reasoning \emph{steps} and the three metrics used to evaluate head-level performance.

\subsection{Defining Reasoning Steps}
To formally quantify head performance during multi-hop reasoning, we define \emph{steps} as the Breadth-First Search (BFS) expansion depth along reasoning paths. This is distinct from the conventional definition of a \emph{hop} (i.e., the final shortest-path length) and allows us to capture head behaviors at each layer of candidate exploration. For example, if the topic entity is ``Barack Obama,'' Step-1 includes triples directly connected to him, such as $(\text{Obama}, \text{born\_in}, \text{Honolulu})$. Step-2 then expands from the entities retrieved in Step-1, considering candidates like $(\text{Honolulu}, \text{located\_in}, \text{Hawaii})$.

\subsection{Head Performance Metrics}
We define three complementary metrics to evaluate the role and effectiveness of each attention head during the reasoning process.

\begin{description}[leftmargin=*, style=sameline]
    \item[\textbf{Contribution}] Measures a head’s share of credit for retrieving correct triples at a given step. It is defined as the proportion of correctly retrieved triples for which this head provided the highest score.
    \begin{equation}\label{eq:contrib}
        \text{Contribution}(h, t) = \frac{|C(h, t) \cap A_t \cap G_t|}{|A_t \cap G_t|}
    \end{equation}

    \item[\textbf{Use Rate}] Measures the model's overall reliance on a head. It is the proportion of samples where a triple scored highest by head $h$ was ultimately selected by the model's gating mechanism at step $t$.
    \begin{equation}\label{eq:userate}
        \text{Use Rate}(h, t) = \frac{|\mathcal{S}(h, t)|}{|\mathcal{S}_t|}
    \end{equation}

    \item[\textbf{Hit Rate}] Measures the precision of a head's top-scoring suggestions. It is the percentage of a head’s suggestions selected by the gate that are actually correct.
    \begin{equation}\label{eq:hitrate}
        \text{Hit Rate}(h, t) = \frac{|C(h, t) \cap A_t \cap G_t|}{|C(h, t) \cap A_t|}
    \end{equation}
\end{description}
\noindent where, for any given head $h$ and reasoning step $t$, $G_t$ is the set of ground-truth triples, and $A_t$ is the set of triples actually retrieved by the gated model. The term $C(h, t)$ represents the set of candidate triples for which head $h$ provided the highest score among all heads. Finally, $\mathcal{S}_t$ is the set of all samples that require reasoning at step $t$, while $\mathcal{S}(h, t)$ is the subset of those samples where a triple from $C(h, t)$ was selected by the final gating mechanism.

\subsection{Head Specialization with Alternative Backbone Encoders}
\label{app:backbone}

To confirm the generalizability of the head specialization phenomenon, we tested the ParallaxRAG framework using two alternative, high-performance backbone encoders: \texttt{intfloat/e5-large-v2} and \texttt{thenlper/gte-large}. Our analysis confirms that the core phenomenon of head specialization persists across all tested encoders, though the specific cooperative patterns among heads vary, reflecting different emergent strategies for multi-hop reasoning.

\subsubsection{\texttt{intfloat/e5-large-v2} Analysis (Figure~\ref{fig:e5})}

The \texttt{e5-large-v2} encoder consistently validates the functional division of labor. Head 10 (Contribution: $0.60$, Use Rate: $0.96$) and Head 2 (Contribution: $0.52$, Use Rate: $0.94$) emerged as the dominant initial-step specialists (peaking at Step-1 for entity localization). Conversely, Head 3 demonstrated the highest precision for terminal reasoning, achieving a Hit Rate of $0.50$ at Step-4. This confirms the learned segregation between high-activation front-end heads and high-precision terminal heads. However, the magnitude of the dynamic switching effect was marginally less pronounced compared to BGE-M3, suggesting a more continuous contribution profile.

\begin{figure*}[h] 
    \centering
    
    \begin{subfigure}[b]{\textwidth} 
        \centering
        \includegraphics[width=1.0\linewidth]{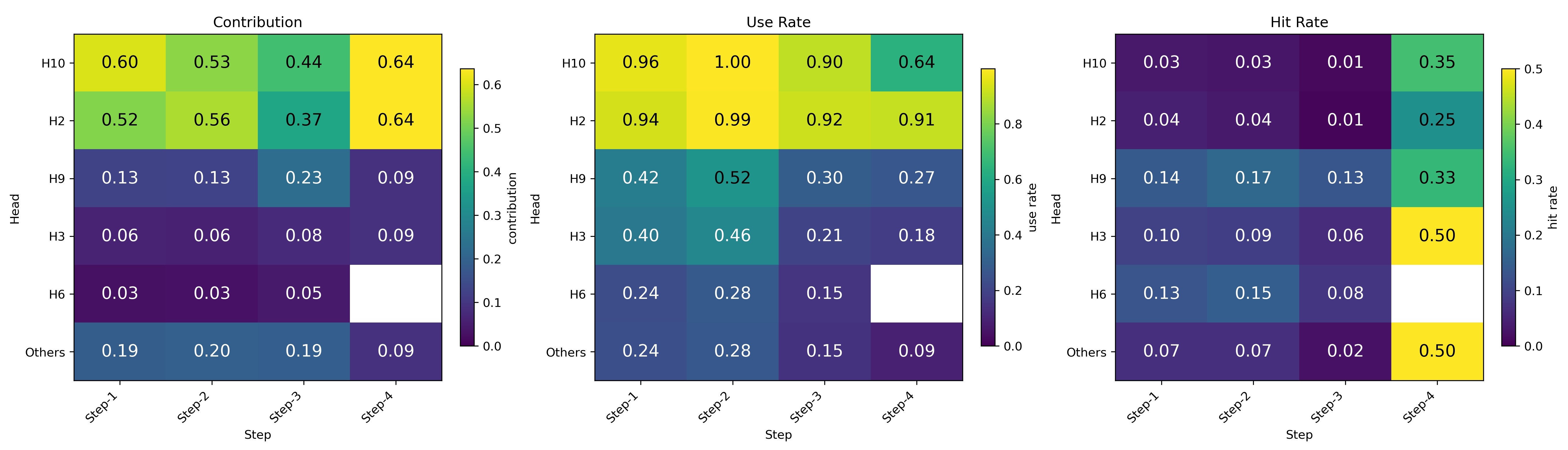} 
        \caption{E5-large-v2 on CWQ dataset}
        \label{fig:e5} 
    \end{subfigure}
    
    \vspace{0.5em} 
    
    \begin{subfigure}[b]{\textwidth} 
        \centering
        \includegraphics[width=1.0\linewidth]{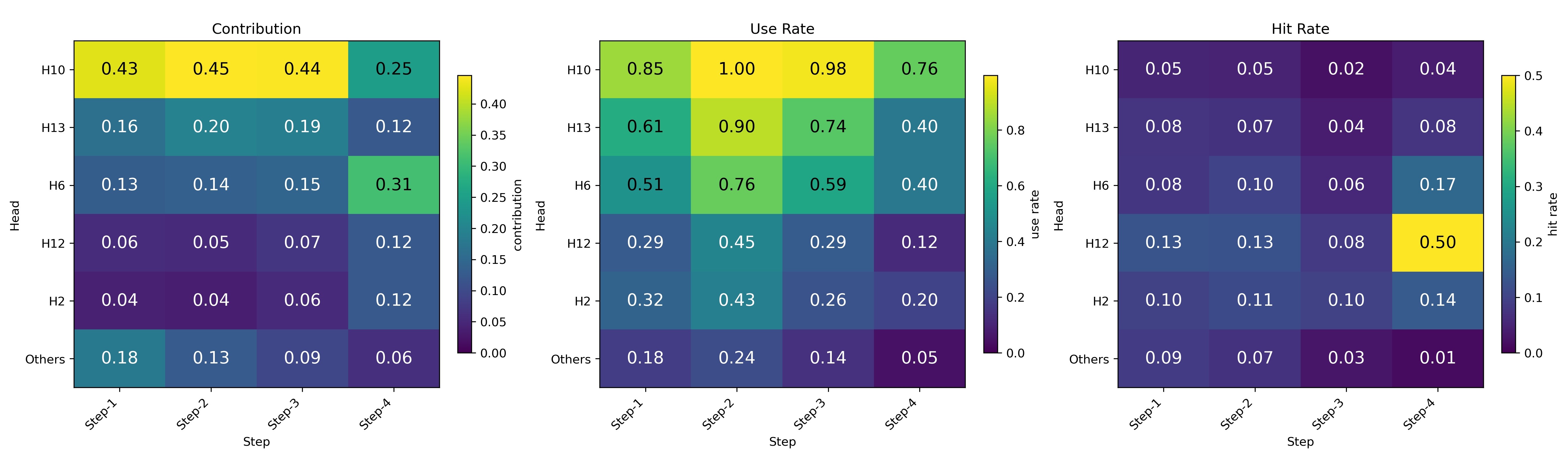}
        \caption{GTE-large on CWQ dataset}
        \label{fig:gte} 
    \end{subfigure}
    
    \caption{Attention Head Specialization Analysis on CWQ Dataset with Alternative Backbone Encoders. (a) Results using the E5-large-v2 model. (b) Results using the GTE-large model.}
    \label{fig:combined_backbone_analysis}
\end{figure*}

\subsubsection{\texttt{thenlper/gte-large} Analysis (Figure~\ref{fig:gte})}

The \texttt{gte-large} results offer a clearer and more compelling validation of dynamic head specialization switching. Head 10 served as the definitive initial-step specialist (Contribution: $0.43$, Use Rate: $0.85$ at Step-1). Crucially, Head 6 and Head 12 collectively assumed the role of long-range dependency specialists in later stages. Head 6's contribution increased significantly at Step-4 (from $0.13$ to $0.31$), marking a clear functional transition. Most notably, Head 12 achieved a maximum Hit Rate of $0.50$ at Step-4, despite minimal involvement at Step-1 (Contribution: $0.06$). This sharp contrast in activation profiles unequivocally demonstrates the learned non-substitutability of specialized heads, confirming that ParallaxRAG successfully engineers a task-adaptive retrieval architecture irrespective of the foundational text encoder.

\begin{table}[h]
    \centering
    \caption{End-to-End QA Performance with Alternative Backbone Encoders using Llama3.1-8B as KGQA generator}
    \label{tab:ener}
    \footnotesize
    \begin{tabular}{lcccc}
        \toprule
        \textbf{Encoder} & \multicolumn{2}{c}{\textbf{WebQSP}} & \multicolumn{2}{c}{\textbf{CWQ}} \\
        \cmidrule(lr){2-3} \cmidrule(lr){4-5}
         & \textbf{Macro-F1} & \textbf{Hit} & \textbf{Macro-F1} & \textbf{Hit} \\
        \midrule
        E5 & 69.23 & 85.63 & 44.59 & 53.95 \\
        GTE & 69.75 & 85.92 & 47.12 & 57.48 \\
        \bottomrule
    \end{tabular}
\end{table}
\subsubsection{Correlation with Performance}
The observed specialization patterns strongly correlate with end-to-end performance, particularly on the long-chain reasoning CWQ benchmark (Table~\ref{tab:ener}). The GTE model, which exhibited a more pronounced dynamic specialization, significantly outperforms the E5 model on CWQ (Macro-F1: $47.12$ vs. $44.59$; Hit Rate: $57.48$ vs. $53.95$). Performance on the simpler WebQSP task is comparable (Macro-F1 difference is $0.52$). This evidence supports the conclusion that the degree of functional specialization learned by the backbone encoder is a critical factor for robustness in multi-hop KGQA.
\subsection{Head Specialization Verification Details}
\label{app:veri}
\subsubsection{Linear Probing}
To quantitatively verify that different attention heads specialize in distinct reasoning stages, we conducted a linear probing experiment. For each reasoning step, we aggregated the output scores from all attention heads to form a feature vector. A logistic regression classifier was then trained on these features to decode the current step number (from 1 to 4). Evaluated under 5-fold cross-validation grouped by sample, the classifier achieved a decoding accuracy that significantly outperformed a random baseline. This result confirms that the activation patterns of the attention heads contain sufficient information to distinguish between different stages of the reasoning process, supporting the hypothesis of functional specialization.
\subsubsection{Difference in Difference in Difference}
To causally validate the functional importance of specialist heads, we designed a Triple-Difference (DDD) intervention. This analysis isolates the additional performance degradation from ablating specialist heads versus random heads on long-hop ($>1$-hop) questions, relative to short-hop (1-hop) questions. We first compute the Difference-in-Differences (DID) for both the specialist-ablation ($\text{DID}_{\text{spec}}$) and random-ablation ($\text{DID}_{\text{rand}}$) scenarios. The final causal estimate is then given by the Triple-Difference: $DDD = \text{DID}_{\text{spec}} - \text{DID}_{\text{rand}}$. Our experiment yielded a statistically significant DDD estimate, confirmed via bootstrap confidence intervals and permutation tests. This provides functional evidence consistent with the view that specialist heads are non-substitutable and disproportionately important for complex, multi-hop reasoning.

\subsubsection{Head Disruption Experiments}
\label{app:disruption}

Full disruption results and analysis are presented in Section~\ref{sec:heads} (Table~\ref{tab:disruption}).

\section{Additional Experiment Results}

\subsection{Statistical Stability Analysis}
\label{app:stats}

To verify that the reported improvements are not due to random variation, we conducted five independent training runs and evaluated on the CWQ test set using Qwen3-30B as the generator.
Table~\ref{tab:stability} reports the mean and standard deviation across runs.
Performance is stable across all metrics, with standard deviations below 2 points, confirming that the gains are consistent and reproducible.

\begin{table}[h]
\centering
\caption{Performance stability across 5 independent runs on WebQSP and CWQ (Qwen3-30B generator). Standard deviations are all below 2.0 points.}
\label{tab:stability}
\footnotesize
\resizebox{\linewidth}{!}{ 
\begin{tabular}{lcccccc}
\toprule
 & \multicolumn{3}{c}{\textbf{WebQSP}} & \multicolumn{3}{c}{\textbf{CWQ}} \\
\cmidrule(lr){2-4} \cmidrule(lr){5-7}
\textbf{Model} & \textbf{Macro-F1} & \textbf{Hit} & \textbf{Hallu.} & \textbf{Macro-F1} & \textbf{Hit} & \textbf{Hallu.} \\
\midrule
ParallaxRAG & 75.97 $\pm$ 0.88 & 92.36 $\pm$ 1.60 & 75.85 $\pm$ 0.63 & 59.15 $\pm$ 1.39 & 66.08 $\pm$ 1.92 & 57.72 $\pm$ 1.01 \\
\bottomrule
\end{tabular}
} 
\end{table}

Additionally, we performed paired bootstrap resampling (1k samples) on Hit@1 over CWQ using Llama3.1-8B as generator.
Improvements over SubgraphRAG ($p = 0.022$) and GNN-RAG ($p = 0.010$) are statistically significant, substantially reducing the likelihood that the observed gains arise from random fluctuations.

\subsection{End-to-End Latency Breakdown}
\label{app:latency}

Table~\ref{tab:latency_webqsp} and Table~\ref{tab:latency_cwq} report a full per-stage latency breakdown (mean $\pm$ std, in seconds per query) on WebQSP and CWQ respectively, measured on a single NVIDIA RTX 6000 Ada GPU.
Retrieval latency increases with hop count due to larger subgraph sizes, but is largely invariant to the retrieval budget $k$ because the multi-view scoring and gating are implemented as fully vectorized matrix operations over the entire candidate triple set.
Consequently, increasing $k$ enriches the LLM context without incurring additional retrieval overhead.
The dominant cost in the pipeline is the LLM generation stage; the multi-view retrieval mechanism itself adds only modest overhead.
\begin{table}[h]
\centering
\caption{Per-stage latency (seconds/query, mean $\pm$ std) on WebQSP. ``Retrieval (mean)'' is averaged over all queries; hop-specific values reflect subgraph complexity.}
\label{tab:latency_webqsp}
\footnotesize
\resizebox{\linewidth}{!}{ 
\begin{tabular}{lcccccc}
\toprule
\textbf{Method} & \textbf{1-hop} & \textbf{2-hop} & \textbf{$\geq$3-hop} & \textbf{Retrieval} & \textbf{KG Access} & \textbf{Generation} \\
\midrule
ParallaxRAG & 22.10 $\pm$ 7.80 & 70.47 $\pm$ 14.28 & -- & 38.78 & 0.023 & 3.124 $\pm$ 0.232 \\
\bottomrule
\end{tabular}
} 
\end{table}

\begin{table}[h]
\centering
\caption{Per-stage latency (seconds/query, mean $\pm$ std) on CWQ.}
\label{tab:latency_cwq}
\footnotesize
\resizebox{\linewidth}{!}{ 
\begin{tabular}{lcccccc}
\toprule
\textbf{Method} & \textbf{1-hop} & \textbf{2-hop} & \textbf{$\geq$3-hop} & \textbf{Retrieval} & \textbf{KG Access} & \textbf{Generation} \\
\midrule
ParallaxRAG & 25.58 $\pm$ 8.20 & 87.95 $\pm$ 17.17 & 183.63 $\pm$ 63.01 & 82.35 & 0.026 & 3.784 $\pm$ 0.227 \\
\bottomrule
\end{tabular}
} 
\end{table}

After extraction, triple recall is computed deterministically via scripted string matching against annotated ground truth, so the final metric does not depend on GPT-4o's generative tendencies.
To assess extraction stability, we repeated the extraction with three different models (GPT-4o, Qwen3-30B, GLM-4.7) and observed negligible variance (std $\leq$ 1.5\%).
We also manually inspected a random 10\% subset and found high consistency between automated extraction and human judgment.


\section{Case Study}
\label{app:case}
We present several representative cases from the CWQ dataset to demonstrate ParallaxRAG’s efficacy in multi-hop question answering, emphasizing its refined subgraph construction, reduced hallucinations, and robust handling of constraints through multi-view decoupling and head specialization, while also revealing scenarios where increased retrieval breadth may be unnecessary for simpler queries.

\begin{figure*}[htbp]
    \centering 
    \includegraphics[width=1\textwidth]{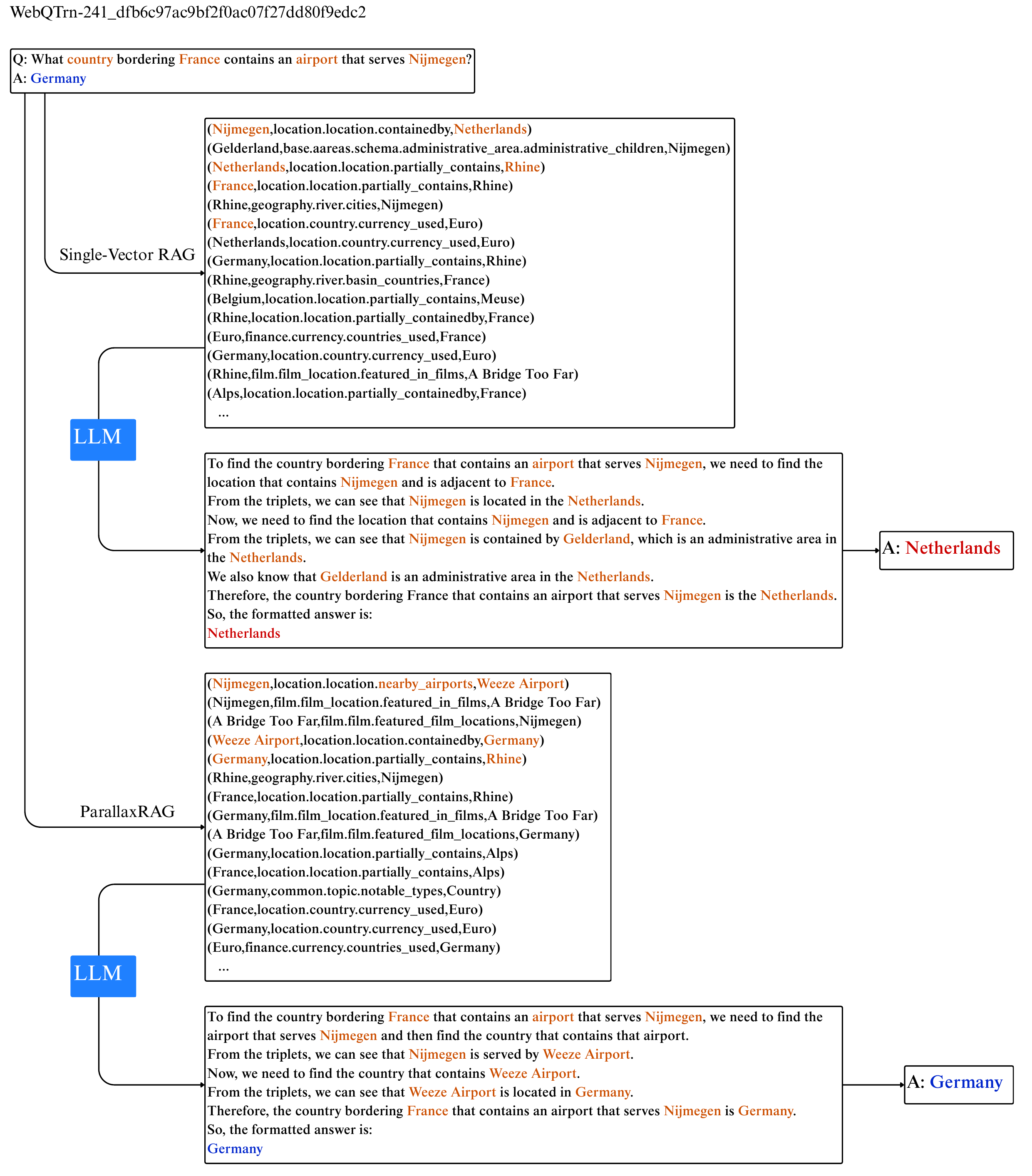} 
    \caption{Comparison of retrieved subgraphs and reasoning chains for case WebQTrn-241 } 
    \label{fig:case1} 
\end{figure*}

\paragraph{Case 1 (Figure~\ref{fig:case1}):}
Single-Vector RAG retrieves a noisy subgraph, incorrectly following the containment path \textit{(Nijmegen, ..., Netherlands)} while ignoring the adjacency constraint to France, thus outputting \texttt{Netherlands}. This failure highlights how its undifferentiated embeddings struggle to enforce multiple constraints. In contrast, ParallaxRAG's multi-view retrieval isolates the correct evidence path: \textit{(Nijmegen, nearby\_airports, Weeze Airport)}, \textit{(Weeze Airport, containedby, Germany)}, and \textit{(Germany, adjoins, France)}. By decoupling and prioritizing geographic constraints, it correctly answers \texttt{Germany}.

\begin{figure*}[htbp]
    \centering 
    \includegraphics[width=1\textwidth]{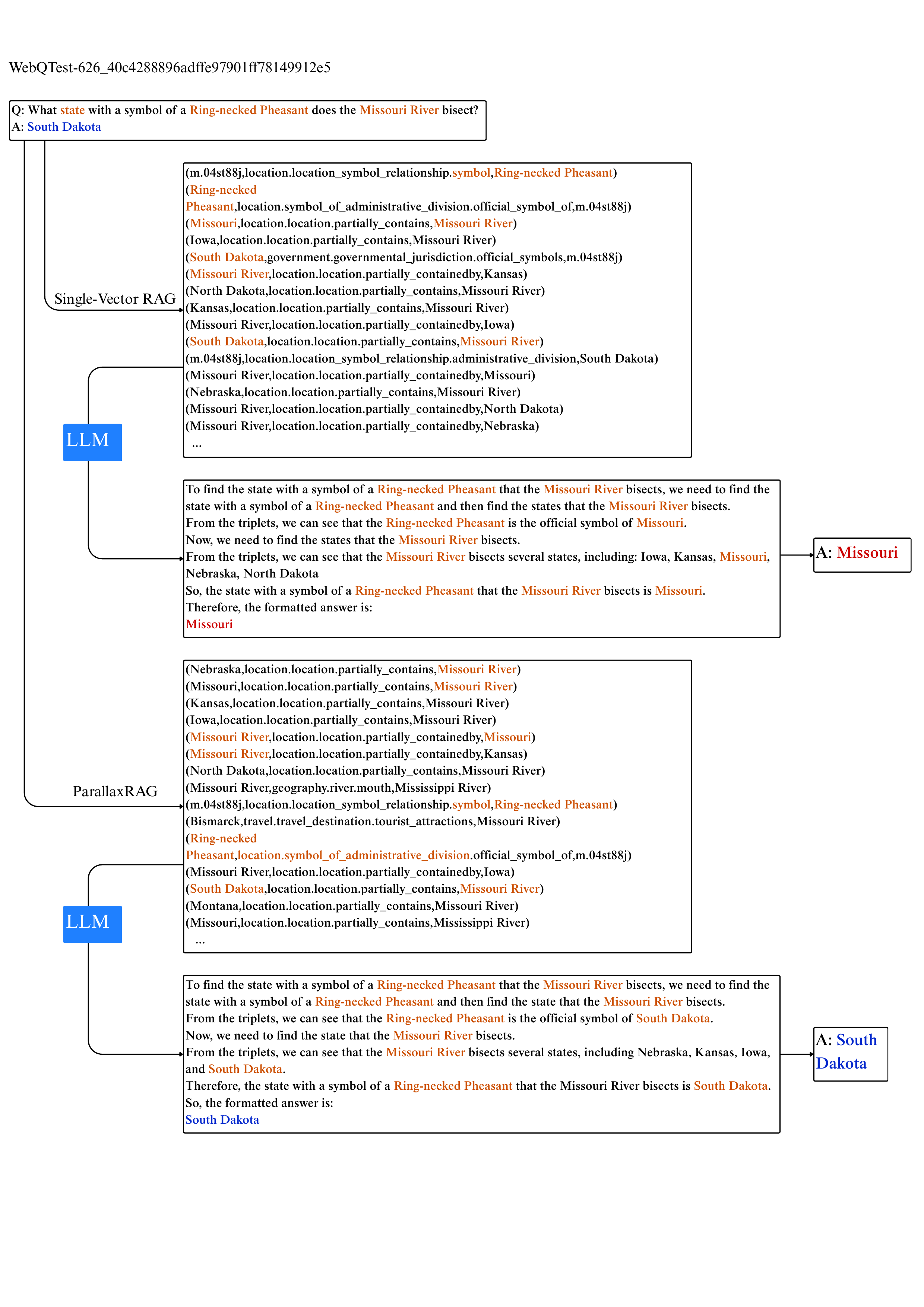} 
    \caption{Comparison of retrieved subgraphs and reasoning chains for case WebQTest-626} 
    \label{fig:case2} 
\end{figure*}

\paragraph{Case 2 (Figure~\ref{fig:case2}):}
Single-Vector RAG's flat representation fails to manage two distinct constraints (symbol and bisection). It retrieves a diffuse subgraph, conflates relations, and incorrectly links the Ring-necked Pheasant to Missouri, thus outputting \texttt{Missouri}. ParallaxRAG, however, uses constraint-specific heads to retrieve precise and separate facts: \textit{(Ring-necked Pheasant, official\_symbol\_of, South Dakota)} and \textit{(South Dakota, partially\_contains, Missouri River)}. This decomposition allows it to satisfy both constraints and correctly identify \texttt{South Dakota}.

\begin{figure*}[htbp]
    \centering 
    \includegraphics[width=1\textwidth]{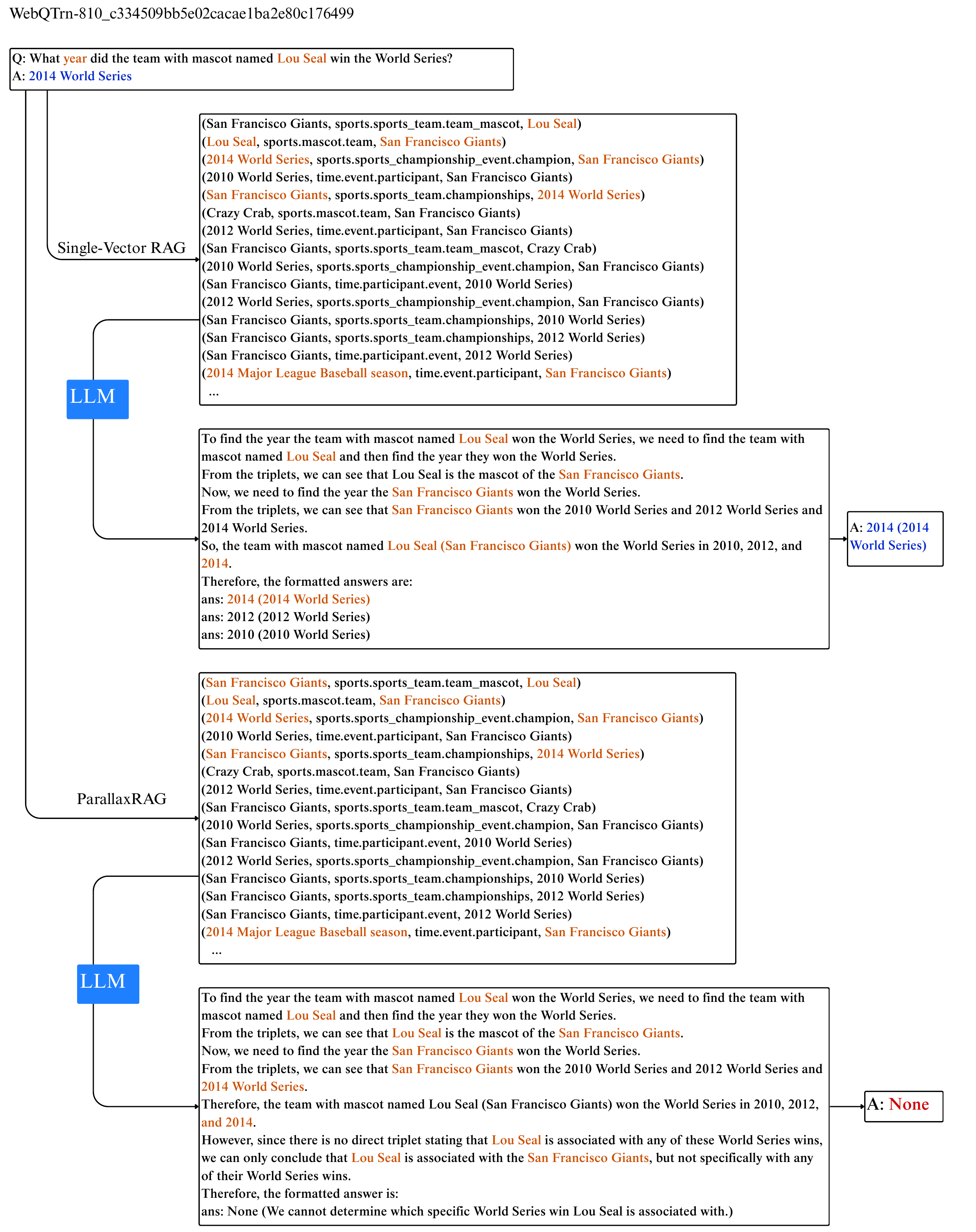} 
    \caption{Comparison of retrieved subgraphs and reasoning chains for case WebQTest-810} 
    \label{fig:case3} 
\end{figure*}

\paragraph{Case 3 (Figure~\ref{fig:case3}):}
This query requires only simple transitive reasoning, where a compact set of facts is sufficient to infer the correct answers. As shown in Figure~\ref{fig:case3}, Single-Vector RAG retrieves a concise collection of relevant triples, enabling the downstream LLM to directly enumerate all championship years. In contrast, ParallaxRAG retrieves a richer set of team- and season-related triples from multiple views. While all retrieved evidence is factually correct, the additional contextual information is not strictly necessary for this shallow inference. Consequently, the downstream LLM adopts a conservative interpretation strategy and outputs an inconclusive answer. This case highlights the scope of multi-view retrieval. While richer contextual coverage is crucial for robustness in complex multi-hop reasoning, ParallaxRAG is not explicitly optimized for minimal-context inference in shallow queries, where simpler retrieval strategies may already suffice.


\end{document}

%% file: ParallaxRag-math_commands.tex
\usepackage{amsmath,amsfonts,bm}

\def\eqref#1{equation~\ref{#1}}

\def\1{\bm{1}}

\DeclareMathAlphabet{\mathsfit}{\encodingdefault}{\sfdefault}{m}{sl}
\SetMathAlphabet{\mathsfit}{bold}{\encodingdefault}{\sfdefault}{bx}{n}

